\documentclass[10pt,conference,a4paper]{IEEEtran}
\IEEEoverridecommandlockouts
\usepackage{cite}
\usepackage{amsmath,amssymb,amsfonts}
\usepackage{algorithmic}
\usepackage{graphicx}
\usepackage{textcomp}
\usepackage{verbatim}
\usepackage{booktabs}
\usepackage{epsfig}
\usepackage{xcolor}
\def\BibTeX{{\rm B\kern-.05em{\sc i\kern-.025em b}\kern-.08em
    T\kern-.1667em\lower.7ex\hbox{E}\kern-.125emX}}
\begin{document}

\title{Text Synopsis Generation for Egocentric Videos}

\author{\IEEEauthorblockN{Aidean Sharghi}
\IEEEauthorblockA{\textit{Department of Computer Science} \\
\textit{University of Central Florida}\\
Orlando, FL \\
aidean.sharghi@gmail.com}
\and
\IEEEauthorblockN{Niels da Vitoria Lobo}
\IEEEauthorblockA{\textit{Department of Computer Science} \\
\textit{University of Central Florida}\\
Orlando, FL \\
niels@cs.ucf.edu}
\and
\IEEEauthorblockN{Mubarak Shah}
\IEEEauthorblockA{\textit{Department of Computer Science} \\
\textit{University of Central Florida}\\
Orlando, FL \\
shah@crcv.ucf.edu}}
\maketitle

\begin{abstract}
Mass utilization of body-worn cameras has led to a huge corpus of available egocentric video. Existing video summarization algorithms can accelerate browsing such videos by selecting (visually) interesting shots from them. Nonetheless, since the system user still has to watch the summary videos, browsing large video databases remain a challenge. Hence, in this work, we propose to generate a \textbf{textual synopsis}, consisting of a few sentences describing the most important events in a long egocentric videos. Users can read the short text to gain insight about the video, and more importantly, efficiently search through the content of a large video database using text queries. Since egocentric videos are long and contain many activities and events, using video-to-text algorithms results in thousands of descriptions, many of which are incorrect. Therefore, we propose a multi-task learning scheme to simultaneously generate descriptions for video segments and summarize the resulting descriptions in an end-to-end fashion. We Input a set of video shots and the network generates a text description for each shot. Next, \textbf{visual-language content matching unit} that is trained with a \textbf{weakly} supervised objective, identifies the correct descriptions. Finally, the last component of our network, called \textbf{purport} network, evaluates the descriptions all together to select the ones containing crucial information. Out of thousands of descriptions generated for the video, a few informative sentences are returned to the user. We validate our framework on the challenging UT Egocentric video dataset, where each video is between 3 to 5 hours long, associated with over 3000 textual descriptions on average. The generated textual summaries, including only 5 percent (or less) of the generated descriptions, are compared to groundtruth summaries in text domain using well-established metrics in natural language processing. \footnote{Research was sponsored by the Army Research Office and was accomplished under Grant Number W911NF-19-1-0356. The views and conclusions contained in this document are those of the authors and should not be interpreted as representing the official policies, either expressed or implied, of the Army Research Office or the U.S. Government. The U.S. Government is authorized to reproduce and distribute reprints for Government purposes notwithstanding any copyright notation herein.}
\end{abstract}

\vspace{-10pt}
\section{Introduction}
\label{intro}
As video acquisition devices become more and more widespread, the corpus of available video content is exponentially growing. With emergence of video logging (vlog) phenomenon and increase in utilizing body-worn and dash cameras by police officers and everyday users, the corpus of long egocentric videos has grown exponentially. Due to the length and volume of these videos, it is extremely challenging to browse, index, or extract information from this rich resource. Hence, effective video summarization frameworks are now necessity rather than luxury.

Traditionally, video summarization techniques aim to create a short video summary that consists of a diverse set of important frames or shots. To infer knowledge about a video, system users can watch the summary video to save time and effort. While this might not impose a problem when the database is small, however, it is extremely challenging and computationally expensive to search through the content of a large database of egocentric videos (or their video summaries).

To tackle the issues mentioned above, in this work, we develop a novel framework to generate {\em text synopsis} for a given {\em egocentric videos}. Similar to conventional methods, the input to our model is a video, however the output in our system is a short textual synopsis that consists of several sentences that are chosen based on their correctness in describing the events in the video as well as their significance in conveying important information about the video when considered together. System users can read the short text to gain insight about the video, and more importantly, users can browse a large number of videos as quickly as possible without watching them. Furthermore, one can conveniently search through a large video database via text queries (text search is by far faster than searching in videos).

One straightforward to generating text synopsis for a video is to first select important/interesting shots in the video using an off-the-shelf video summarization algorithm. Having selected a set of shots, one can feed each shot to a pre-trained video-to-text network, also known as video caption generator, to generate a description for it. By putting all the generated descriptions into a document, one obtains the textual synopsis associated with the video. 

There are two major flaws in models following this architecture. Firstly, since the video summarization algorithm merely considers the low-level visual features of a video shot to measure its interestingness, it may select shots that are visually different, but represent the similar activity or event. Similar descriptions are generated by the captioning network for such similar shots, hence, resulting in a textual summary with redundancy of information. Secondly, even if the video summarizer chooses perfect shots, generating correct descriptions for long egocentric videos is extremely challenging. Therefore, if not dealt with, some descriptions included in the textual summary of the video could be wrong. These flaws result in suboptimal text synopses. We report performance in section~\ref{sec:exp} for this kind of approach.

Another approach to generating text synopsis for a video is to first generate dense captions using existing video-to-text algorithms to obtain a long text describing all events in the video, important or not. Next, a text summarization model can be employed to summarize this long document by selecting a few most representative sentences. The major flaw in such models originates from the disconnect between the captioning network and the text summarization unit. The captioning module will generate many descriptions (thousands per video in our experiments), some of which may be meaningless or incorrect. However, the text summarization algorithm works under the premise that the given document is minimally noisy. As we will demonstrate in section~\ref{sec:exp}, this assumption does not hold, especially when dealing with long egocentric videos.

To resolve the issues mentioned above, we propose a multi-task training scheme that joins the two tasks of caption generation and text summarization in an end-to-end fashion. This eliminates the disconnect between the two tasks and allows the model to both effectively densely caption the video and summarize the large pool of captions, taking into consideration that some may be wrong. To achieve this goal, first visual features of the video shots are fed to an LSTM-based video caption generation network to generate a textual description for each shot, in form of a natural language sentence. 
Next, each generated sentence is passed through a module that assigns to it a correctness score that measures how well it describes the visual content of the video shot. In parallel, another module processes all the generated sentences in temporal order, assigning a significance score to each based on their importance in the context of the video. Subsequently, summary-level impact of each sentence, represented by a scalar, is formulated as the product of its correctness and significance scores. This way the worthiness of a sentence based on its individual quality as well as the importance of the event that it is describing in the context of the video is evaluated. The summary-level impact values in the temporal order form a time series where the peaks correspond to locally important events in the video. Finally, the sentences with peak impact values are collected and this text synopsis is returned to the user.

We evaluate our model on UT Egocentric~\cite{lee2012discovering}, a challenging video summarization dataset that consists of long videos (3 to 5 hours each) acquired using head-mounted cameras. These videos contain several daily life activities (e.g. driving, shopping, dining, etc.) in uncontrolled environments. While it is extremely challenging to summarize such videos, it is where video summarization techniques are most helpful. Textual summaries generated by our model are compared to groundtruth summaries in text domain via ROUGE~\cite{lin2004rouge} metric, a well-established metric in automatic document summary evaluation. Our method outperforms the baselines with a significant margin, achieving the state-of-the-art performance.

To the best of our knowledge, we are the first to formulate video text synopsis generation via an \textbf{end-to-end multi-task learning} scheme. Furthermore, since it is practically infeasible to acquire annotations on whether every description generated by the captioning network during training epochs is correct or not, we design an efficient \textbf{weakly} supervised objective to automatically identify correct descriptions. There are three main advantageous to our model: 1) our model allows the users to browse a large number of videos as quickly as possible without watching them, 2) it also enables exploiting semantic information in its most familiar form, i.e. in natural language domain, to conveniently search through a large video database via text queries, 3) in addition to the textual synopsis, video shots corresponding to the sentences in the synopsis can be easily retrieved and put together to generate a visual summary similar to the output of conventional methods.

\section{Related Work}

Traditionally, video summarization techniques aim to create a short video summary that consists of a diverse set of important frames or shots. Early works used low-level appearance or motion feature descriptors~\cite{goldman2006schematic,laganiere2008video,liu2002optimization,rav2006making,wolf1996key,zhao2014quasi}, whereas recent works take advantage of more sophisticated features to enhance the quality of generated video summaries~\cite{gong2014diverse,kim2014joint,kwon2012unified,xiong2014detecting,yaohighlight,zhang2016summary}. Even though summaries created by these techniques are shorter than the original videos, system users must watch these summary videos to extract information. This problem can be alleviated by generating textual synopses for the videos. It is rather easy to read a textual summary about a video that consists of a few sentences, and more importantly, users can search through the content of many videos efficiently using text queries.

Generating text descriptions \textbf{(not summaries)} for videos is also known as video caption generation. In \textbf{single-sentence captioning}, as the name suggests, the objective is to produce a single description for a video. Early efforts such as~\cite{guadarrama2013youtube2text,rohrbach2013translating}, are template-based, that is to fill in part-of-speech tags with the actions, places, and objects. After the re-emergence of neural networks, due to their superior performance, several methods following an encoder-decoder structure were proposed~\cite{chen2017generating,gan2017semantic,pan2016jointly,venugopalan2015sequence}. Such frameworks use recurrent neural networks such as LSTM to first encode the video, and then generate a caption for it using a decoder. More recent works take use attention mechanism to boost the performance~\cite{xu2015show}.

\begin{figure*}[t]
	\centering
	\includegraphics[width=0.9\linewidth]{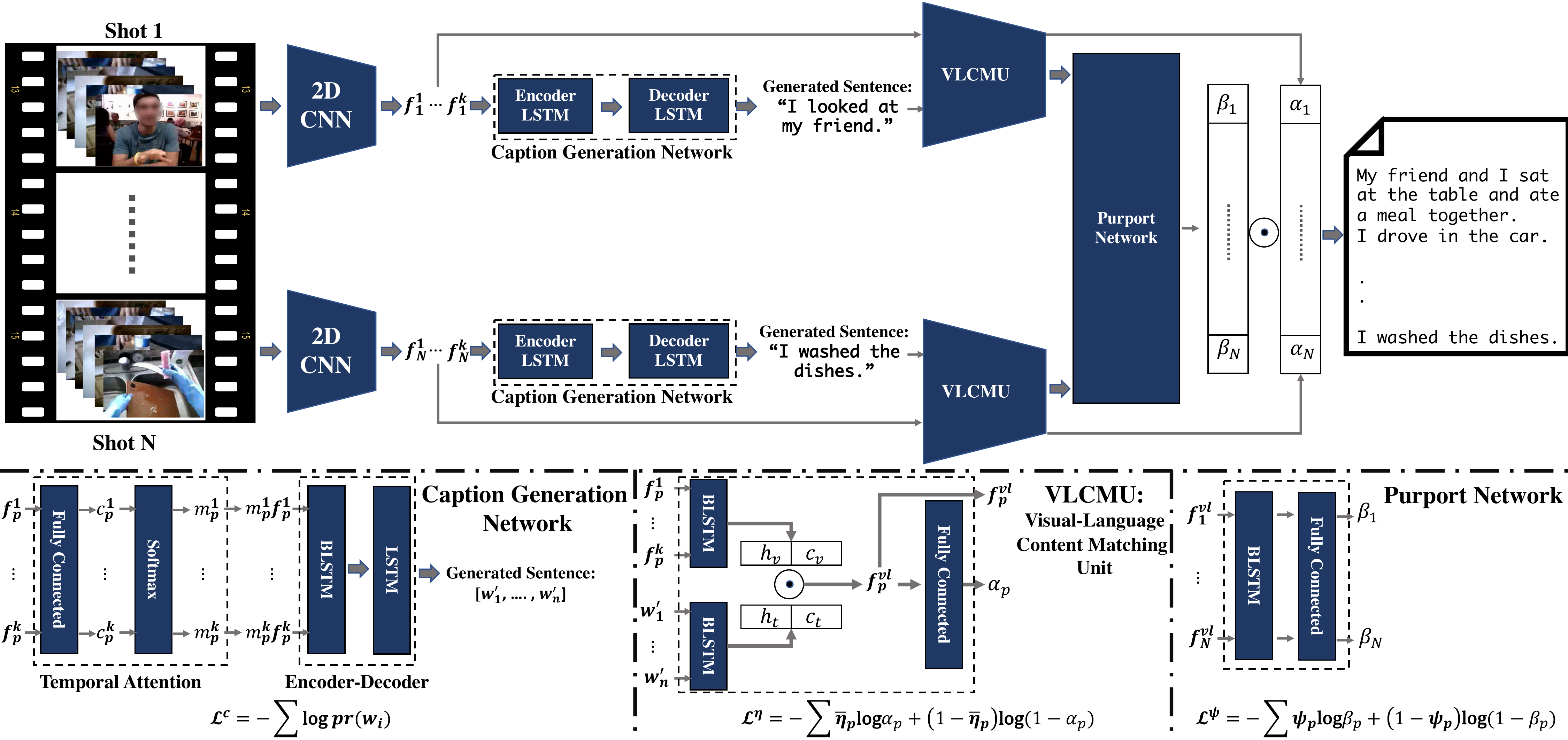}
	\caption{A complete overview of our proposed approach. Given a video, we partition it uniformly into shots, subsample $k$ frames from each shot, and pass them to a convolutional neural network for feature extraction (\{$\textbf{\textit{f}}_p^{1},\cdots,\textbf{\textit{f}}_p^{k}$\}, where $p$ indicates the shot index). The set of features for each shot is then fed to a caption generation network that produces a sentence describing it. Our Visual-language Content Matching Unit (VLCMU) learns to assign a scalar value $\alpha_p \in [0,1]$ to each description based on how correct it is in describing the video shot. In addition to $\alpha_p$, VLCMU produces a \textit{visual-language} feature representation, $\textbf{\textit{f}}_p^{vl}$, for a shot given its visual features and generated description. The feature representations of all shots in the video are then passed through our purport network. Unlike VLCMU that works on shot level, purport network performs at video level, taking all visual-language shot representations to assign a a scalar value $\beta_p \in [0,1]$ to each, measuring importance of $p^{th}$ shot's \textbf{description} in the context of the video. Finally, impact score of a description, $\gamma_p$, is computed as the product of its correctness and significance scores, $\alpha_p\times\beta_p$. The descriptions pertaining to peak significance scores are put in the text synopsis and are returned to the user.} 
	\label{fig:app} 
	\vspace{-12pt}
\end{figure*}

The goal in \textbf{dense video captioning}~\cite{wang2018bidirectional,zhou2018end} is to produce multiple captions. In such models, a proposal module selects segments from the video, and a description for each segment is generated by the captioning network. From the pool of all generated captions, those with highest confidence (that is a function of both proposal accuracy and caption quality) are returned to the user. Due to challenging nature of generating dense captions for untrimmed videos that contain several activities, normally, many descriptions are generated, some of which may be incorrect. Existing works are evaluated on ActivityNet dataset~\cite{krishna2017dense} that includes 20K videos, each 120 seconds long on average.

There are two key differences between textual summarization of egocentric videos and dense video captioning. Firstly, the latter aims to produce as many captions as possible whereas in the former task, we aim to return a few most representative descriptions that capture the important events and activities occurring in the video. Secondly, in text summary generation for videos, it is crucial we identify the correct descriptions and select the most informative descriptions among the correct ones. Specially when dealing with long egocentric videos, we must note that dense captioning will result in a few thousand descriptions, many of which are either meaningless or incorrect in describing the video segment. We verify this claim by generating dense captions for the videos and using a SoTA text summarization algorithm to summarize this long text document. As we show in the experiment section of this paper, such approaches will fail to produce satisfactory results. This is indeed because existing text summarization algorithms do not account for severe noise in the text document.

In \textbf{video captioning with a paragraph}~\cite{yu2016video,xiong2018move}, model generates a detailed description of the video by generating few ``semantically-fluent'' sentences. While we share spirit with works of this genre to a certain extent, the existing approaches work well for short videos (few minutes long).

One can feed the output of dense video captioning to a \textbf{document summarization} algorithm~\cite{carbonell1998use,cheng2016neural,edmundson1969new,mcdonald2007study,chopra2016abstractive,pighin2014modelling,thadani2013supervised} to generate a short text summary. According to~\cite{radev2002introduction}, a summary is text produced from one or more text documents that preserves the important information of the original texts and is usually significantly shorter than half of the original documents. As mentioned earlier in the introduction, such algorithms work when the given document is minimally noisy. In section~\ref{sec:exp}, we experimentally confirm this claim.

To the best of our knowledge, Sah et al.~\cite{sah2017semantic} is the only existing work that tackles text synopsis generation for egocentric videos. Their model first visually summarize the videos by selecting the interesting segments, followed by a pre-trained captioning network that generates descriptions per segment, and finally a text summarization algorithm that summarizes the pool of descriptions. As explained earlier, due to the disconnect between the individual components (video summarization, caption generation, and text summarization), the generated synopses are suboptimal. 

\section{Approach}
\label{approach}
Given a long egocentric video, our main goal is to produce a textual summary or report of it that consists of sentences. To this end, we propose an end-to-end multi-task learning scheme to simultaneously densely caption the video and summarize the large pool of generated captions. These tasks, when learned together, allows us to identify which video shots do not yield correct captions, hence significantly boosting the chance of including correct descriptions in the eventual textual summary. We explain components of our model in details in the remainder of this section.

\subsection{Unified text synopsis generation framework}
We illustrate our complete framework for producing textual synopsis from videos in Figure~\ref{fig:app}. Given a video, first we partition it uniformly into non-overlapping shots due to three main reasons. First, the UT Egocentric dataset employed to validate our model on is annotated the same way. Yeung et al.~\cite{yeung2014videoset} uniformly partitioned each video in UT Egocentric dataset into non-overlapping 5-sec long shots and collected a textual description for each. This allows us to effectively train and evaluate our model. Second, since almost all events in the videos are longer than shot length, splitting the events into short shots results in multiple visually similar clips that have same or similar descriptions. This allows us to effectively train our caption generation network (we need more than one sample per description to train a network that can generalize). Third, this is indeed the most efficient way to densely caption a long egocentric video. Any alternative approach requires either a proposal module (to propose video segments) or a shot boundary detector, both of which yield unnecessary complications in the model. It is also worth mentioning that even if the captioning network produces identical descriptions for consecutive shots, this does not reduce the performance. This is due to the fact that we remove duplicate sentences that appear consecutively during inference, all events (whether long or short) is described in a single short sentence. Hence, uniformly splitting the videos does not pose a problem.

Next, $k$ frames are sampled from each shot's pool of frames. Each frame is then passed through a pre-trained CNN for feature extraction. Thus, each shot is represented by the set of its frame-level feature representations, i.e. \{$\textbf{\textit{f}}_p^{1},\cdots,\textbf{\textit{f}}_p^{k}$\}, where $p \in \{1,\cdots,N\}$ indicates the shot index number. The rest of our pipeline consists of: 1) a caption generation network, 2) a visual-language matching unit, and 3) a purport network. In the following, we discuss the details of these submodules.
\subsubsection{Video caption generation network}
Given a shot's feature set, we wish to generate a sentence that describes it. Since each shot is a short video clip, we develop our own caption generation network, instead of using an off-the-shelf model. The reason behind is that the complexity of these models mainly originate from a proposal unit that is unnecessary due to short length of the shots. To confirm this claim, we adopted the captioning network of Wang et al.'s~\cite{wang2018bidirectional} and performed two experiments. In our first experiment, we initialized their pre-trained (on ActivityNet Captions~\cite{krishna2017dense}) model, and fine-tuned it on UT Egocentric~\cite{lee2012discovering}. This yielded significantly lower performance. The underlying reason behind such inferior performance is the significant vocabulary difference between the two datasets. In other words, existing captioning datasets are not suitable for captioning long egocentric videos and will not provide improvements. In a second experiment, we trained their model from scratch on UT Egocentric, and were only able to achieve comparable results. Therefore, we proceeded with our own design.

Detailed structure of our caption generation network is presented in bottom left corner of Figure~\ref{fig:app}. It consists of a temporal attention module, a bidirectional LSTM that serves as the encoder, and a decoder LSTM that generates the sentences.

More formally, given a feature set, \{$\textbf{\textit{f}}_p^{1},\cdots,\textbf{\textit{f}}_p^{k}$\}, of a shot, each $\textbf{\textit{f}}_p^{i}$ is fed to fully connected layer with output dimension of 1, i.e. a scalar output $c_p^{i}$. A softmax activation is applied on top of [$c_p^{1},\cdots,c_p^{k}$] to transform them into probability values [$m_p^{1},\cdots,m_p^{k}$], that serve as weights for the original features. The purpose of the designated temporal attention module is to find most representative features of a shot and only use those for caption generation. This is a crucial step, specially when dealing with egocentric videos in which the camera is constantly moving. The temporal attention module discards uninformative feature representations to increase the quality of generated sentences.

We feed the weighted features to the encoder, that is a bidirectional LSTM, and obtain the cell and hidden states. These states are then used to initialize the decoder states which produces a sentence for each shot. Conventional sum of negative log likelihood of the words is used as the objective function to train our caption generation network: 
\begin{align}
\vspace{-5pt}
{\cal L}^c_p = -\sum_{i=1}^m \log(pr(w_i)),
\label{eq:1}
\vspace{-5pt}
\end{align}
where $m$ is the number of words in the groundtruth sentence, $w_i$ is the $i^{\text{th}}$ word in it and $pr(.)$ yields the likelihood of the input word. Note that ${\cal L}^c_p$ in Equation~\ref{eq:1} is for $p^{\text{th}}$ shot and the overall caption generation loss (${\cal L}^c$) is computed by summing loss of all captions in training set.

\subsubsection{Visual-Language Content Matching Unit}
\label{sec:nat}
A well-trained caption generation network generates sentences that resemble sentences written by human subjects. For a long egocentric video, we generate a few \textbf{thousand} descriptions, however, not all of them are descriptive of their corresponding shots (i.e. the sentence is meaningful on its own but it is a wrong description for the given video clip). In fact, our study shows that only a small fraction of the generated sentences are informative. Thus, it is crucial that our model can distinguish between informative/uninformative sentences. 

To recognize that a generated sentence is a wrong description for a given shot, only attending to the sequence of words in the sentence is no longer sufficient. These sentences resemble those written by human subjects, but do not describe the scene. To identify such cases, we must attend to both the visual features as well as the sequence of words in the generated sentence.

Formally, our Visual-Language Content Matching Unit (VLCMU) takes as input visual features \{$\textbf{\textit{f}}^1_p,\cdots,\textbf{\textit{f}}^k_p$\} of a shot, as well as its corresponding generated sentence \{$\textbf{\textit{w}}^{'}_{1},\cdots,\textbf{\textit{w}}^{'}_n$\} (where $n$ is the number of words in the sentence and $w^{'}_i$ is the $i^{\text{th}}$ word in $p^{\text{th}}$ shot's sentence), and assigns to it a scalar score $\alpha_p \in [0,1]$. Note that we omit subscript $p$ in notation used for words in the sentence to enhance legibility. Ideally, this unit assigns scores close to $0$ to uninformative and $1$ to informative sentences respectively.

Our complete VLCMU network is shown in Figure~\ref{fig:app} at the bottom. Two bi-directional LSTM units are employed, one processing the visual features of a shot and the other reads its corresponding generated sentence word-by-word. Hidden and cell states of each BLSTM unit are obtained and concatenated. Resulting representations are then elementwise multiplied to produce a \textit{visual-language} feature representation, $\textbf{\textit{f}}_p^{vl}$, for that shot. Indeed visual and textual features belong to different feature spaces, however, LSTM
32 has been proven sufficient in learning the common embedding as well as processing the sequences at once~\cite{duan2018weakly}. Finally, we feed this new feature representation to a fully connected layer with sigmoid nonlinearity to obtain the correctness score $\alpha_p$.
\begin{table*}[t]
	\centering
	\footnotesize
	\caption{Comparison results between our proposed approached and several state-of-the-art video summarization algorithms that are adapted to produce textual synopsis. All numbers reported below are F1 scores reported by the metric used.}
	\label{tab:res}
	\begin{tabular}{@{}lcccccccc@{}}\toprule
		& \multicolumn{1}{c}{SeqDPP\cite{gong2014diverse}} &
		\multicolumn{1}{c}{Superframes\cite{gygli2014creating}}& 
		\multicolumn{1}{c}{SubMod\cite{gygli2015video}}&
		\multicolumn{1}{c}{$\text{SubMod}_{\text{VL}}$\cite{plummer2017enhancing}} &
		\multicolumn{1}{c}{LM-SeqGDPP\cite{sharghi2018improving}} &
		\multicolumn{1}{c}{LSTMDPP\cite{zhang2016summary}} & 
		\multicolumn{1}{c}{\textbf{Ours}}\\ 
        \midrule
        ROUGESU4 & 11.95 & 15.59 & 16.19 & 14.10 & 16.24 & 13.98 & \textbf{17.33}\\
        ROUGE-L   & 11.41 & 13.88 & 16.24 & 13.95 & 16.99 &  12.84 & \textbf{24.98}\\
        METEOR & 19.02 & 18.36 & 18.90 & 18.02 & \textbf{19.90} & 17.32 & 19.27\\
        BLEU2 & 16.69 & 34.68 & 35.28 & 35.38 & 35.11 & 31.72 & \textbf{46.90}\\
		\bottomrule
	\end{tabular}
	\vspace{-10pt}
\end{table*}
We use a binary-crossentropy loss ${\cal L}^\eta$ for this network that allows us to learn its parameters. For a generated sentences, we set its \textbf{pseudo groundtruth} correctness score $\bar{\eta}_p$ to 1 if more than half of its constituent words appear in the groundtruth sentence and otherwise to 0. This \textbf{weak} supervision effectively eliminates the need to annotate the generated sentences during every epoch of training. ${\cal L}^\eta$ is computed as follows:
\vspace{-5pt}
\begin{align}
{\cal L}^\eta = -\sum_{p=1}^N\bar{\eta}_p \log \alpha_p + (1 - \bar{\eta}_p) \log (1-\alpha_p),
\end{align}
As shown in our ablation study (please refer to Table~\ref{tab:ab}), the existence of this unit is crucial to the performance of the model. This confirms our intuition that without identifying correct descriptions, the summaries will suffer quality loss. This is simply because out of thousands of descriptions generated by the captioning network, less than 5 percent are selected to be put in the final text synopsis.

\subsubsection{Purport network}
So far, our model is able to 1) generate a sentence, and 2) assign a correctness score to it for every shot in the video. In other words, in the pool of sentences generated for the entire video, our model is able to identify the informative sentences. The final step in the pipeline is to make a coherent text synopsis for the video by choosing the sentences with highest summary-level impact. 

To achieve this, we propose to learn it directly from user summaries. Hence, given the visual-language features of all shots, $\{\textbf{\textit{f}}_1^{vl},\cdots,\textbf{\textit{f}}_N^{vl}\}$, we pass them through a bidirectional LSTM to obtain get their forward and backward representations, followed by a fully connected layer with sigmoid nonlinearity to obtain a shot-level significance score $\beta_p$. 

Using a binary-crossentropy loss, we can learn Purport network's parameters:
\begin{align}
\vspace{-5pt}
{\cal L}^\varphi = -\sum_{p=1}^N\varphi_p \log \beta_p + (1 - \varphi_p) \log (1-\beta_p),
\label{eq:pur}
\vspace{-5pt}
\end{align}
where $\varphi_p$ is set to 1 for the shots whose corresponding groundtruth sentences are present in the user summary and it is set to 0 otherwise.
It is worth noting that while the purport network is used to essentially summarize the pool of sentences by selecting informative ones, it is \textbf{not} a text summarization module. Text summarization models assume that the given document's constituent sentences are correct. However, as we study in section~\ref{sec:exp}, a majority of the generated sentences (at least for long egocentric videos) are wrong in describing their corresponding shots. Hence, the purport network uses the visual-language features produced by VLCMU that is responsible in identifying such faults in the generated sentences.

\subsubsection{Training and inference}
To train the model presented in Figure~\ref{fig:app}, first we pre-train the caption generation network. Once it is trained, we freeze its weights to train the rest of the network. The overall objective function of the full pipeline is:
\begin{align}
\vspace{-5pt}
{\cal L} = \sum_{v=1}^V{\cal L}_v^c + \lambda_1 {\cal L}_v^\eta + \lambda_2 {\cal L}_v^\varphi,
\label{eq:l}
\vspace{-5pt}
\end{align}

where $V$ is the number of training videos and $\lambda_1$ and $\lambda_2$ are hyper-parameters that adjust the weights for each term in the overall objective function. At the test time, model generates a sentence for each shot in the video and assigns a correctness and a significance score ($\alpha_p$ and $\beta_p$ respectively) to each sentence. Summary-level impact value $\gamma_p$ of a shot is considered as the product of its correctness and significance. These impact values in the temporal order form a time series, where the peaks correspond to locally (in time) important events in the video. While fairly simple and intuitive, there are two main advantages to using such test time inference algorithm. Firstly, since the peaks are local maxima (as opposed to simply choosing sentences with highest $\gamma$ values), the produced textual synopsis is uniform in time. Secondly, by repeating the inference process multiple times on the $\gamma$ time-series, one can produce shorter and shorter textual summaries. Each time, we make a smaller time series by only keeping the peaks in previous one. It is easy to infer that text summary is guaranteed to get truncated with a minimum rate of $\frac{1}{2}$ each time the inference algorithm is applied (in worst case, every other point in the time series is a peak, and hence the algorithm discards the non-peak point).

Moreover, after the text synopsis is generated, we can optionally retrieve their corresponding video shots, put them together in temporal order to create a visual synopsis, simulating the standard video summarization functionality.
\begin{table*}[t]
	\centering
	\footnotesize{
	\caption{We compare the visual synopsis using AUC metric introduced by \cite{sharghi2018improving}.}
	\vspace{-5pt}
	\label{tab:res2}
	\begin{tabular}{@{}lcccccccc@{}}\toprule
		& SeqDPP~\cite{gong2014diverse} & Superframes~\cite{gygli2014creating} & SubMod~\cite{gygli2015video} & $\text{SubMod}_{\text{VL}}$~\cite{plummer2017enhancing} & LM-SeqGDPP~\cite{sharghi2018improving} & LSTMDPP~\cite{zhang2016summary} &
		\textbf{Ours}\\ 
        \midrule
        AUC   & 12.91 & 9.85 & 11.53 & 12.26 &  13.14 & 7.57 & \textbf{13.19}\\
		\bottomrule
	\end{tabular}
	\vspace{-7pt}		}
\end{table*}

\section{Experimental Setup and Results}
\label{sec:exp}
\textbf{Dataset.} Several video summarization datasets~\cite{de2011vsumm,gygli2014creating,song2015tvsum} exist, however, most are not readily adaptable for textual video summarization. Furthermore, many consist of short videos (up to 10 minutes long). Video summarization is in fact most helpful when the videos are long and carry high degree of redundancy. Hence, following~\cite{lu2013story,lee2015predicting}, we train and test our framework (and baselines) on long egocentric videos that Lee et al~\cite{lee2012discovering} collected. This dataset consists of 4 videos, each between 3$\sim$5 hours long, covering a variety of daily life activities such as driving, shopping, dining, studying, etc. in uncontrolled environments. Yeung et al.~\cite{yeung2014videoset} uniformly partitioned each of these videos into non-overlapping 5-second long shots and collected a single sentence description for each via Amazon Mechanical Turk. These descriptions serve as groundtruth in training our caption generation network, and also to infer the pseudo groundtruth labels $\bar{\eta}_p$ used in training our VLCMU.

For any shot index $p$ that is specified to be important $\varphi_p$ is 1, otherwise it is set to 0. This allows us to learn our purport network parameters via Equation~\ref{eq:pur}.

\textbf{Train/test splits.} Given that there are four videos in UT Egocentric dataset~\cite{lee2012discovering}, we run four rounds of experiment. Each time, we leave one of the videos out for testing and use the remaining three in training phase. The pool of shots in the training videos are split again to 80 and 20 percent for training and validation respectively to train the caption generation module. After that the training videos are used to train the remaining parts of the network. The trained model is then applied on the test video to obtain its text synopsis.

\textbf{Features.} We sub-sampled 6 frames from each shot. We use a pretrained GoogLeNet~\cite{szegedy2015going} network for the 2D CNN block in Figure~\ref{fig:app} to extract a 1024-d visual feature representation for each sampled frame. These representations serve as our shot-level features \{$\textbf{\textit{f}}_p^{1},\cdots,\textbf{\textit{f}}_p^{6}$\}. 

\textbf{Details of inference algorithm.} Our inference algorithm is designed such that the user is able to obtain summaries of different granularity. Therefore, each time we run the inference algorithm to further refine the summary for a less detailed summary. In other words, the peaks that were selected at the previous run, are fed as the new input to the inference algorithm. In our experiments, we run the inference algorithm 4 times for all videos. Indeed, running the inference multiple times is rather necessary. The captioning network generates on average about 3000 descriptions for each video (this makes sense because the videos are 3 to 5 hours long). However, conventionally it is expected for video summarization models to generate summaries that are 5 percent (or less) of the overall length of the videos. Following this standard, by running the inference algorithm multiple times, we discard 95 percent of the descriptions to keep the most representative descriptions.


\textbf{Evaluating text synopses.} Since existing video summarization frameworks work in the video domain (i.e. they produce a visual synopsis as opposed to a textual summary), they are not readily comparable to our model. To overcome this problem, we propose to accompany these frameworks with a video caption generation network. In other words, we adapt them to produce textual synopsis for a video. Given a video, each baseline summarizes the video by selecting its key shots. These shots are then passed one by one through caption generation network, identical to that employed in our model (bottom left corner of Figure~\ref{fig:app}), that produces a sentence describing each shot. All sentences of selected key shots are pooled together to construct the textual summary. This resembles how Sah et al.~\cite{sah2017semantic} tackled the problem of text synopsis generation for videos. In this work, we use more complex summarization methods that outperform their model. Now that the baselines are adapted to produce textual synopsis for videos, we can compare all the models using existing NLP-based metrics. Yeung et al.~\cite{yeung2014videoset} were the first to evaluate video summaries through text. More specifically, they used ROUGE-SU, one of the multiple metrics offered by the ROUGE~\cite{lin2004rouge} evaluation toolbox. To ensure a more comprehensive study, we also report results using ROUGE-L, METEOR~\cite{banerjee2005meteor}, and BLEU2~\cite{papineni2002bleu}, common metrics in evaluation of machine translation.

\textbf{Evaluating visual summaries.} As mentioned earlier, after a text synopsis is generated for the video, we can retrieve their corresponding video shots, put them together in temporal order to create a visual synopsis. The visual summaries are directly comparable to summaries generated by a conventional video summarization method. Therefore, we compare our models against state-of-the-art methods in visual domain. To do so, we follow \cite{sharghi2018improving} by using dense concept annotations to compare the system summaries with their corresponding reference summaries. 

\begin{figure*}[t]
\centering
\includegraphics[width=0.85\linewidth]{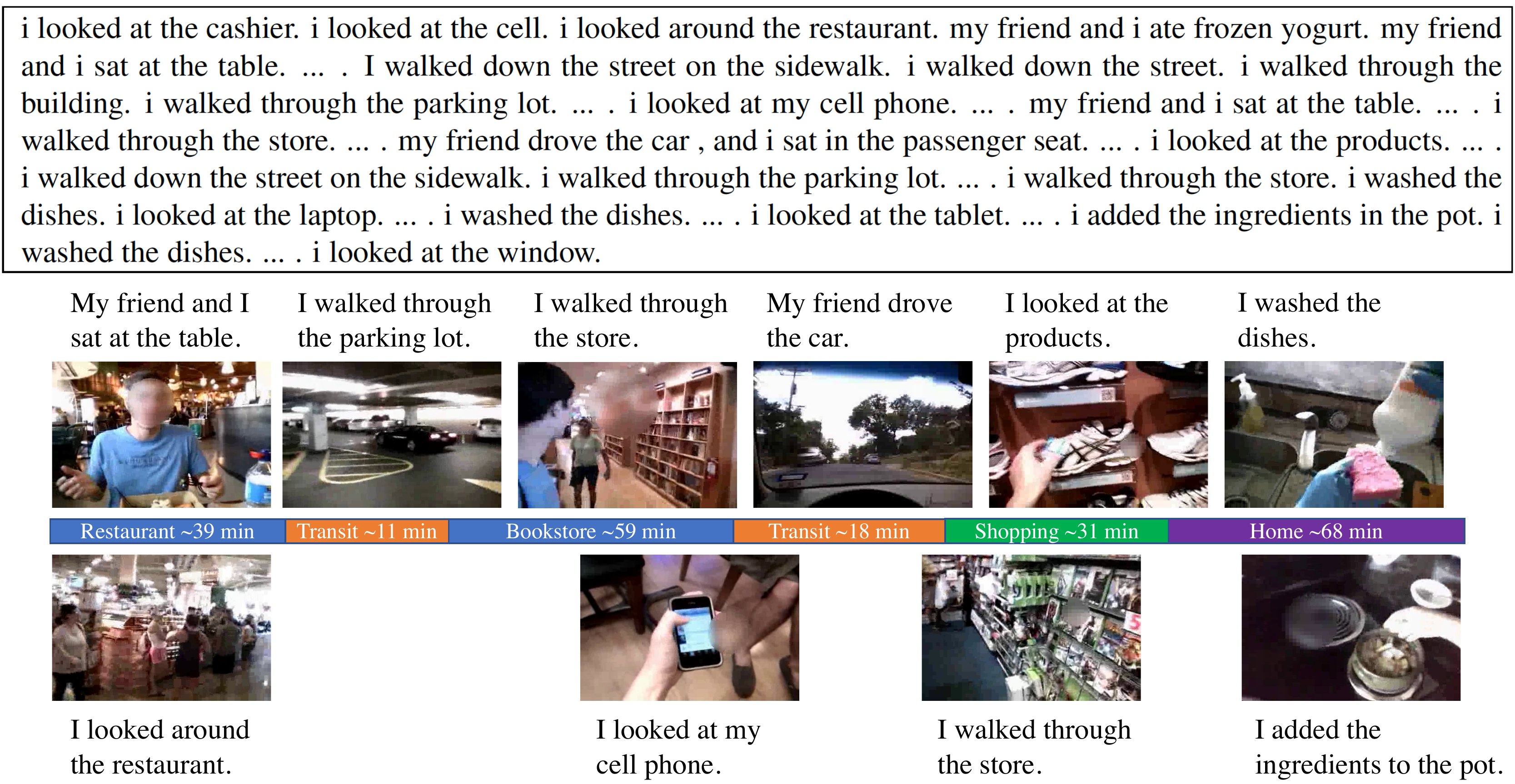}
\vspace{-7pt}
\small{\caption{Generated textual synopsis for video 1 in the UT Egocentric dataset is illustrated below. Densely captioning of this video leads to obtaining over 3600 descriptions, out of which our model selected 144 descriptions for the final text synopsis of the video (at the top, we only show few sentences for brevity purposes. At the bottom, we show some keyframes associated with the descriptions generated.}}
\vspace{-10pt}
\label{fig:p01}
\end{figure*}

\textbf{Results analysis.} As mentioned earlier, for each video in the dataset, three groundtruth text summaries are available. Each system summary is compared to all three reference summaries. Therefore, we obtain three F1 scores (using the precision and recall reported by ROUGE-SU4) for each system generated summary and report their average. We compare our model with several video summarization methods listed above in Table~\ref{tab:res}. We are able to achieve state-of-the-art performance on every video in UT Egocentric dataset. As we can observe from the table, SubMod~\cite{gygli2015video} and LM-SeqGDPP~\cite{sharghi2018improving} have relatively higher performance compared to the other approaches. This is in fact because both of these methods are fed the expected summary length. In other words, they produce system summaries with exact number of sentences that of in their corresponding groundtruth summaries. Since F1 score is the harmonic mean of precision and recall, it leads to an advantage for these methods. It is also worth noting that even though LSTMDPP~\cite{zhang2016video} also has this advantage, it is unable to perform well. This is because the summaries produced by this approach are \textit{not} uniform in time and repetition can be observed in their summaries. SeqDPP by~\cite{gong2014diverse} produced long summaries compared to other methods as it has no mechanism for controlling the summary length. This leads to its fairly low performance in our experiments. In case of Video 3 and 4, the summaries produced by our method were roughly 50 percent shorter than the groundtruth summaries, and even so, we outperform all other baselines. Only the summary generated for Video 1 was slightly longer than its groundtruth. Figure~\ref{fig:p01} illustrates a sample summary from our model.

We also evaluated \textbf{visual} synopses under the AUC metric of \cite{sharghi2018improving}. As shown in Table~\ref{tab:res2}, we outperform the existing state-of-the-art. This shows that transferring the summarization task from the visual domain to text domain results in generating better summaries in \textbf{both} domains.

\textbf{Ablation study.} To illustrate the effectiveness of the introduced VLCMU and purport network, we remove each from the pipeline with minimal changes to the remaining structure. As we explained earlier, the purpose of our VLCMU is to identify which generated captions are informative. Hence, we expect to observe a noticeable drop in performance when we eliminate this module. When this module is entirely removed (i.e. including the BLSTM units that process visual features and generated text), we notice a performance drop of 1.43 percent in the F1 score. However, we can keep the BLSTM units and only remove the auxiliary loss ${\cal L}^\eta$ (and the corresponding fully connected layer that produces $\alpha_p$ that is designed to enhance learning its parameters. When ${\cal L}^\eta$ is removed from the objective function, we observe an insignificant drop in performance. This is in fact because we use pseudo-groundtruth labels to calculate ${\cal L}^\eta$ to avoid extra annotation. In another experiment, we remove the purport network. In this experiment, the pseudo groundtruth labels $\bar{\eta}_p$ in ${\cal L}^\eta$ are replaced with $\varphi_p$. In this case, we observe that the performance drops 0.81 percent in F1 score. This is expected as we are only considering each generated on its own and independent of all other sentences. 

Finally, we design an experiment in which we remove both the VLCMU and purport network from the pipeline, and instead apply a document summarization method to obtain the textual summary. To do this, we generate the descriptions for all shots in a video and put them together in a text document. This document is then summarized using the state-of-the-art extractive document summarization method of~\cite{dong2018banditsum}. This yields a low performance of 12.41 under ROUGE-SU4 F1-score. This is expected as such approaches do not account for the fact that some of the generated descriptions may be incorrect. We summarize these observations in Table~\ref{tab:ab}.
\begin{table}[t]
	\centering
	\caption{Ablation study. We study the effect of removing components, specified by column title, from the pipeline.}
	\vspace{-5pt}
	\label{tab:ab}
	\begin{tabular}{@{}lcccc@{}}\toprule
		         & -VLCMU & -${\cal L}^\eta$ & -Purport & \small{\cite{dong2018banditsum}}\\
		\midrule
        R-SU4 & 15.90 & 17.12 & 16.52 & 12.41 \\
		\bottomrule
	\end{tabular}
\vspace{-15pt}
\end{table}

\section{Conclusion}
In this work, we present a framework that given a video, it produces a short textual synopsis for it. To this end, the video is divided into shots and a descriptive sentence for each shot is generated via a video caption generation network. Since some of the generated sentences may be uninformative specially when dealing with long egocentric videos, we develop a visual-language content matching unit that can distinguish between the informative and uninformative sentences. Next, our purport network reads the generated sentences in temporal order, to select those that convey the most information about the video. To the best of our knowledge, we are the first to develop a unified approach to generating text synopsis for videos by modeling the problem as dense text generation followed by text summarization under severe noise. Our framework can also generate a visual synopsis for the video by retrieving the shots whose corresponding descriptions were included in the textual synopsis and stitch them together in temporal order to make a short video. We evaluate our model on a challenging egocentric dataset where videos are over 3 hours long, and achieve SoTA performance.


\end{document}